\documentclass{article}

\usepackage{arxiv}

\usepackage[utf8]{inputenc} 
\usepackage[T1]{fontenc}    
\usepackage{hyperref}       
\usepackage{url}            
\usepackage{booktabs}       
\usepackage{amsfonts}       
\usepackage{nicefrac}       
\usepackage{microtype}      
\usepackage{lipsum}		
\usepackage{graphicx}
\usepackage{doi}

\usepackage[square, numbers]{natbib}
\usepackage{amsmath}
\usepackage{amssymb} 

\usepackage[ruled,linesnumbered]{algorithm2e}

\usepackage{subcaption} 
\usepackage{csquotes} 

\usepackage{fancyhdr}
\usepackage{lipsum}

\usepackage{hyperref}

\title{Learning to schedule job-shop problems: Representation and policy learning using graph neural network and reinforcement learning}

\author{\href{https://orcid.org/0000-0002-6778-7632}{\includegraphics[scale=0.06]{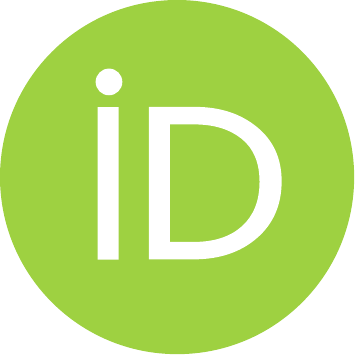}\hspace{1mm}} Junyoung Park\\
	KAIST\\
	Daejeon, Korea\\
	\texttt{junyoungpark@kaist.ac.kr} \\
	\And
	Jaehyeong Chun\\
	KAIST\\
	Daejeon, Korea\\
	\texttt{wogud485@kaist.ac.kr} \\
	\And
	Sang Hun Kim \\
	Samsung Electronics \\
	Suwon, Gyeonggi-do, Korea \\
	\texttt{phd.kim@samsung.com} \\
	\And
	Youngkook Kim \\
	Samsung Electronics \\
	Suwon, Gyeonggi-do, Korea \\
	\texttt{youngkook@samsung.com} \\
	\And
	\href{https://orcid.org/0000-0003-2620-1479}{\includegraphics[scale=0.06]{orcid.pdf}\hspace{1mm}} Jinkyoo Park\\
	KAIST\\
	Daejeon, Korea\\
	\texttt{jinkyoo.park@kaist.ac.kr} \\
}

\date{}

\renewcommand{\headeright}{}

\renewcommand{\shorttitle}{Learning to schedule Job-shop problems with GNN-RL}

\hypersetup{
colorlinks=true,
linkcolor=blue,
filecolor=blue,
urlcolor=blue,
pdftitle={GNN-RL-JSSP},
pdfauthor={Junyoung Park, Jaehyung Chun},
pdfkeywords={Job shop scheduling problem, Graph neural network, Reinforcement Learning},
}

\begin{document}

\renewcommand\footnotemark{}
\thanks{This manuscript is a reprinted version of \href{https://www.tandfonline.com/doi/full/10.1080/00207543.2020.1870013}{the journal manuscript} of same contents.}
\maketitle

\begin{abstract}
We propose a framework to learn to schedule a job-shop problem (JSSP) using a graph neural network (GNN) and reinforcement learning (RL). We formulate the scheduling process of JSSP as a sequential decision-making problem with graph representation of the state to consider the structure of JSSP. In solving the formulated problem, the proposed framework employs a GNN to learn that node features that embed the spatial structure of the JSSP represented as a graph (representation learning) and derive the optimum scheduling policy that maps the embedded node features to the best scheduling action (policy learning). We employ Proximal Policy Optimization (PPO) based RL strategy to train these two modules in an end-to-end fashion. We empirically demonstrate that the GNN scheduler, due to its superb generalization capability, outperforms practically favored dispatching rules and RL-based schedulers on various benchmark JSSP. We also confirmed that the proposed framework learns a transferable scheduling policy that can be employed to schedule a completely new JSSP (in terms of size and parameters) without further training.
\end{abstract}

\keywords{Scheduling \and 
          Job shop scheduling \and
          Job shop scheduling problem \and 
          JSSP \and 
          Graph neural network \and 
          GNN \and 
          Reinforcement Learning \and
          RL}

\section{Introduction}
Scheduling is considered as an important problem in both academia and practice. The job shop scheduling problem (JSSP) is, a type of scheduling problem that aims to determine the optimal sequential assignments of machines to multiple jobs consisting of series of operations while preserving the problem constraints (processing precedence and machine-sharing). The JSSP has attracted much interest in the research community\, not only because it can be utilized as a framework to derive an effective scheduling strategy for manufacturing systems, but also because it can be used as a challenging benchmark for NP-hard combinatorial optimization problems \cite{garey1976complexity} in developing combinatorial optimization (CO) algorithms.

Researchers has suggested numerous approaches to solve the JSSP. Exact optimization approaches such as mathematical programming \cite{manne1960job} and the branch and bound algorithm \cite{lomnicki1965branch}, have been proposed to solve the JSSP. Such methods guarantee an optimal solution but suffer from the curse of dimensionality, which has prevented these methods from being widely applied in the real-world applications \cite{caserta2009metaheuristics}. This limitation has led researchers to focus on developing approximation methods, which aim to find an effective but not optimal solution within reasonable computational time. These methods mainly include neighborhood search methods and priority dispatching rules (PDRs). Neighborhood search methods, such as Tabu search \cite{geiger1997tabu}, simulated annealing \cite{yim1999scheduling}, and genetic algorithm \cite{gupta2006job}, generate high quality solutions in reasonable computational time. However, it is difficult to apply these search-based methods to the dynamic scheduling problems, where the conditions of the problem continuously change, because they are designed to find the entire scheduling assignments for a given initial condition. Therefore, these methods should be applied again whenever the conditions of the scheduling problem change \cite{baykasouglu2017solving}, which is impractical in practice due to the extensive time required to find a good solution.

To cope with such difficulties, PDRs have been extensively applied to real-world manufacturing scheduling problems \cite{panwalkar1977survey, holthaus1997efficient, tay2007designing}. A PDR is a heuristic rule that assigns jobs to machines based on their priorities while considering the current status of the system. By repeatedly applying PDRs whenever the status of the system changes, one can determine a sequence of jobs as the scheduling progress. Due to their simplicity, human-interpretable results, and low computational load, the majority of industries rely on PDRs more than other optimization methods. However, PDRs ignore the sequential nature of scheduling problems. As an extension of PDR, the composite dispatching rules (CDR) has been investigated for solving dynamic JSSP. the CDR whose parameters (the weight of dispatching rules) are optimized with genetic programming methods can yield a qualified dispatching policy. \cite{ozturk2019extracting, tay2008evolving}


Besides the approaches mentioned earlier, machine learning approaches, which especially utilize the function approximation property of the neural network, have been widely investigated to solve the JSSP in supervised learning (SL) setting \cite{willems1995implementing, weckman2008neural, chen2001competitive}. The objective of the SL methods is to learn parameters of the neural network such that the trained networks can replicate the behavior of pre-existing scheduling methods such as mathematical optimizations, search methods, PDRs, or CDRs. On the other hand, reinforcement learning (RL) approaches have been employed to determine the scheduling actions of JSSP without utilizing any existing methods. In RL approaches, a scheduler (often RL agent) learns a dispatching policy that maps the current state of the manufacturing system to the scheduling action while considering the sequential nature of the problem. Following such characteristics, RL approaches successfully produce a scheduling solution of JSSP \cite{aydin2000dynamic,gabel2012distributed,gabel2008adaptive,lin2019smart}. Moreover, the RL approaches often discover the dispatching rules that are superior to the existing PDRs \cite{gabel2012distributed, gabel2008adaptive}. However, these RL methods typically require instance-by-instance training to derive the schedule, which entails difficulty applying the trained policy to a new JSSP.

There have been few studies that employ graph neural networks (GNN) to solve scheduling problems, such as traveling salesman problem (TSP), vehicle routing problems (VRP) \cite{li2018combinatorial, kaempfer2018learning, prates2019learning}. These studies first represent a problem instance into a graph and employ GNN to transform the graph into a set of node embedding that summarizes the structural information of the target problem. The policy function then takes the node embedding and produces the scheduling action. Because GNN enables the scheduling policy to handle the differently-sized graph, the trained policy can be used to produce scheduling actions for problem instances in different sizes.

In this paper, we propose a framework to construct the scheduling policy for JSSPs using GNN and RL. We formulate the scheduling of a JSSP as a sequential decision making problem in a computationally efficient way. Next, we represent the state of a JSSP as a graph, consisting of operation nodes and constraint edges. Then, we employ a GNN to learn node features that embed the spatial structure of the JSSP and derive a scheduling policy that maps the embedded node features to scheduling action. 

We represent the state of job shop using a disjunctive graph \cite{roy1964problemes,yamada2003studies}. The disjunctive graph specifies the structure of JSSP instance; nodes represent operations, conjunctive edges represent precedence/succeeding constraints between two nodes, and disjunctive edges represent machine-sharing constraints between two operations. The graph representation is then processed by GNN, a type of neural network that learns the interaction among nodes and edges of given graphs, to extract the node embedding. Because the node embedding summarizes the structural information of the JSSP graph, the scheduling policy that takes the node embedding as an input can yield good scheduling action while considering the JSSP structure. Since GNN learns how to extract the node embedding from the input graph, the GNN-based JSSP scheduler shows better scheduling performance than well-known PDRs on unseen JSSP instances without additional training. Moreover, the GNN-based scheduler shows on-par or even better scheduling performances than the baseline RL methods that are trained for solving specific-size JSSP problems, again without additional training.

\begin{figure}[t]
  \centering
  \includegraphics[width=0.75\textwidth]{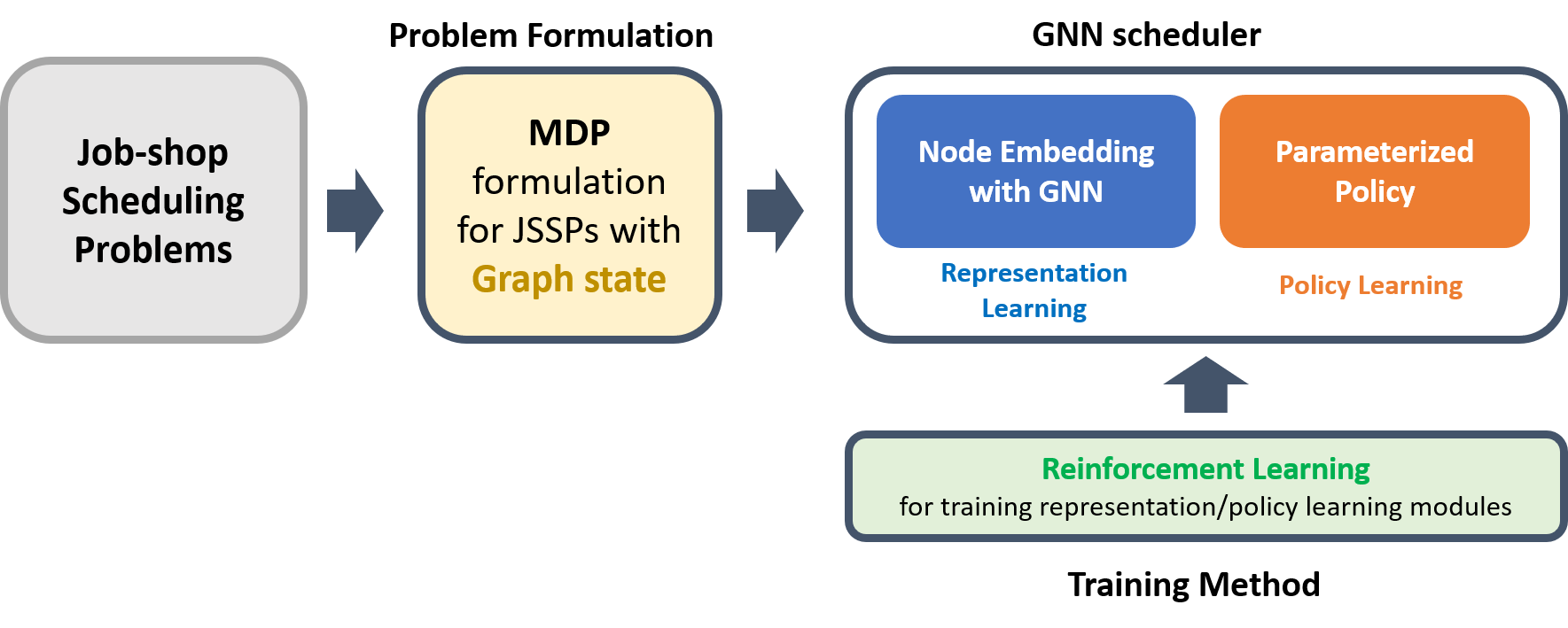}
  \caption{\textbf{An overview of the proposed framework}}
  \label{fig:Overview}
\end{figure}

We use proximal policy optimization (PPO) \cite{schulman2015trust} algorithm, a variant of policy-based RL, to train the GNN based state representation module and the parameterized decision-making policy jointly. PPO updates the parameters of the GNN and the policy only when the current representation module and the policy (locally) improves the scheduling performance. Because PPO have shown the most stable learning performance comparing to other RL algorithms such as baseline-augmented policy gradients \cite{sutton2018reinforcement} and Q-learning \citep{mnih2013playing}, this study employs PPO as principal learning algorithm. In summary, we propose a framework that effectively represents the JSSP state (representation learning) and makes the optimal decision (policy learning), along with an RL-based training method to optimally train these modules in an end-to-end fashion. We will refer to the derived scheduler as `GNN scheduler'. The proposed framework is illustrated in Figure~\ref{fig:Overview}.

The scheduling policy learns to consider the inherent structure of the JSSP, rather than learning the characteristics of a specific JSSP instance. Therefore, the trained scheduling policy can be directly adopted to new JSSP instances without further training process. We empirically show that the GNN scheduler outperforms existing learning-based methods and PDRs when scheduling for unseen JSSP instances. The contributions of this paper are summarized as follows:

\begin{itemize}
    \item We propose a problem formulation for JSSP as a sequential decision making problem with graph representation of the state to consider the structure of JSSP.
    \item We propose a framework to process JSSP graph state into node embedding using a GNN (representation learning) and make scheduling action using the constructed embedding (policy learning).
    \item We propose effective training methods for both (1) GNN based node embedding and (2) parameterized policy using reinforcement learning in an end-to-end fashion.
    \item We empirically demonstrate that the GNN scheduler for JSSP outperforms practically favored dispatching rules on various benchmark JSSP instances and provides an effective scheduling solution to completely new JSSP instances that are too large to be solved with conventional search-based algorithms.
\end{itemize}

\section{Backgrounds}
\subsection{Markov decision process and reinforcement learning}
Markov decision process (MDP) is a tuple $(\cal{S}, \cal{A}, \mathrm{P}, \mathrm{R}, \gamma)$, where the state space $\mathcal{S}$ and action space $\mathcal{A}$ denote the set of states and actions, respectively; the transition probability function $\mathrm{P} : \mathcal{S} \times \mathcal{A} \times \mathcal{S}  \rightarrow [0,1]$ represents the probability of the next state $s_{t+1} \in \cal{S}$ given the current state $s_t \in \cal{S}$ and action $a_t \in \cal{A}$, i.e., $\mathrm{P}(s,a,s') = \mathbb{P}(s_{t+1}=s'|s_t=s, a_t=a)$; the reward function $\mathrm{R} : \mathcal{S} \times \mathcal{A} \times \mathcal{S} \rightarrow [r_{min}, r_{max}]$ describes the immediate reward received after transition from $s$ to $s'$ due to action $a$; and $\gamma \in (0,1]$ is the discount factor. RL considers a sequential decision-making problem as MDP and solves the MDP by optimizing the Bellman equations through iterative learning procedures. The objective of an RL agent is to learn a policy $\pi(a_t|s_t)$ that maximizes the expected cumulative sum of future rewards, namely state-value $V^{\pi}(s_t)$. The definition of state-value function is as follows:
\begin{equation}
\label{eqn:value}
V^{\pi}(s_t) = \mathbb{E}_{a_k \sim \pi}\left[\sum_{k=0}^{T} \gamma^k R_{t+k}(s_k, a_k, s_k'))\right]
\end{equation}
The state-value function computes the expected cumulative rewards when the target system starts from the current state $s_t$ and is operated using the sequential actions selected by the policy $\pi(a_t|s_t)$. The optimal policy $\pi^*(a_t|s_t)$ is the policy that maximizes $V^{\pi}(s)$ for all $s$.

\subsection{Representation learning on graph structured data}
Complexly structured data can be efficiently represented as a graph. A graph is defined as set of nodes and edges. Nodes represent entities of data, and edges represent relationship between the nodes. By utilizing node features and their relationship captured by edges, representation learning on a graph aims to extract optimum features, called embeddings, for various decision making tasks.

A GNN is a type of neural network that takes a graph as input and typically produces an updated graph whose node (or edge) features are embedded with the consideration of edge connectivities and neighborhood node features. A GNN is mainly composed of three distinctive modules: one for learning pair-wise interactions between nodes; module for aggregating interactions and producing an aggregated feature vector, one module for updating node features with consideration of the aggregated feature \cite{battaglia2018relational}. Here, the interaction module and node updating module are typically neural networks.

For updating a target node, a GNN utilizes the interaction modules to compute the pair-wise messages between the target node and its neighborhood nodes. It then utilizes the aggregation module to form a single vector from the received messages. Finally, it utilizes the updating module to produce the updated node embedding using the node feature and the aggregated messages as module input. By leveraging the learning capability of neural networks, GNN achieves state-of-the-art (SOTA) performance across various domains such as per-node regression tasks, graph clustering tasks, language modelling, and control problems \cite{park2019physics, xinyi2018capsule, sanchez2018graph, yang2018glomo}.

By considering graph structure and the associated node features, GNNs achieves not only SOTA performances but also learns generalizable node representations. Due to the pair-wise interaction module, a trained GNN can be applied to the news graphs whose structures (the number of nodes in graph and the edge connectivities) are different from the graphs used for training. As a result, GNN models can successfully perform their tasks without additional training in various fields including robot control, wind power estimation, and city-scale weather prediction.\cite{park2019physics, wang2018nervenet, seo2019differentiable}. We utilize such transferability of GNNs to accomplish the transferable property in our JSSP scheduler.

\subsection{Deep learning for combinatorial optimizations}
Deep learning (DL) approaches have shown promising results for combinatorial optimization (CO) problems. DL approaches often employ a neural network to represent the structure and characteristics of to the CO problem itself or a sub-stage of the problem in the hidden embedding and estimate the solutions of the CO problem using another neural network that utilizes the embeddings. The DL approaches for solving CO can be classified in two ways depending on the training scheme: supervised learning, and RL.

In supervised learning settings, a neural network learns a direct mapping from inputs, the CO problem itself or a sub-stage of the CO problem, to the pre-computed solution of the target problem. This approaches can generate high-quality solutions, comparable even to the mathematical optimization methods, across various CO problems \cite{li2018combinatorial, kaempfer2018learning, prates2019learning}. However, due to their supervised nature, these methods require pre-computed optimal solutions to train the networks. This requirement makes supervised learning less applicable to large scale CO problems.

RL approaches overcome the limitation of the supervised setting by formulating the iterative solution-search process of CO problems as an MDP. In the MDP formulation, the objective of RL methods is to learn the policy that generates the sub-solution on the current stage. By repeatedly applying the policy mapping of the current stage (state) of a target problem to the updated partial solution, RL methods can generate a solution of CO problem. To learn an optimal policy represented as a parametric function, RL methods utilize transition samples (often composed of sub-stage, sub-solution, immediate reward, and the next sub-stage), which can be easily collected through numerical simulations. Although RL methods only utilizes the transition samples during learning, they can successfully solve various CO problems without employing pre-computed optimal solutions \cite{mittal2019learning, kool2018attention, khalil2017learning}.

\section{Problem formulation}
\subsection{Graph state representation}
The disjunctive graph $g = (\cal{V},\cal{C} \cup \cal{D})$ has been used as an effective representation for the JSSP \cite{roy1964problemes}. The operation set $\mathcal{V}$ contains the job-shop's operations which are represented as nodes in the graph. The conjunctive edge set $\mathcal{C}$ contains the conjunctive edges, each of which denotes the precedence constraint between two consecutive operations on the same job. The disjunctive edge set $\mathcal{D}$ contains the disjunctive edges, each of which represents the machine-sharing constraint between two nodes, i.e., when two operations can be processed by the same machine, the corresponding operation nodes are connected with a disjunctive edge.

The disjunctive graph only contains static information of the JSSP such as processing times of operations, precedence sharing, and machine sharing constraints, failing to incorporate dynamically changing JSSP state information. We incorporate node features into the nodes in $g$ so that $g$ represent the current state of the JSSP during scheduling. The node feature $x_v$ assigned to node $v$ is as follows:

\begin{itemize}
    \item Node status: one-hot encoded indicator of the operation status. [1, 0, 0], [0, 1, 0], and [0, 0, 1] indicate the operation is not scheduled yet, is being processed by a machine, and is finished, respectively.
    \item Processing time: the processing time of the operation
    \item Degree of completion: the degree of completion when the operation $v$ is completed, e.g., the degree of completion is 0.4 for the first operation and 0.6 for the second operation in the job composed of two operations which have 20 and 30 unit processing times, respectively.
    \item Number of succeeding operations: 	the number of successor operations in the job that include the operation $v$
    \item Waiting time: the waiting time of the operation $v$ only after the operation is ready to process. 
    \item Remaining time: the remaining time of the operation to be completed. It has value $-1$ if the operation is not scheduled yet or already finished.
\end{itemize}

We refer to the conjunctive graph with the node features as the graph state for the JSSP. We ignore the dummy nodes (source and sink nodes) of the disjunctive graph because they represent imaginary operations and thus have no features in our graph formulation. 

\subsection{MDP formulation of JSSP}
We formulate a scheduling process of a JSSP instance as an MDP. The proposed MDP is a tuple $(\cal{G}, \mathcal{A}, \mathrm{P}, \mathrm{R}, \gamma)$. The state $g_t \in \cal{G}$ denotes a snapshot of the job shop at transition $t$. we assume that $g_t$ contains all information of the JSSP including static features and dynamic features of JSSP which results from scheduling decisions. Note that the transition index $t$ does not coincide with the simulation time stamps. The action $a_t$ is a scheduling action of either (1) loading an operation to an available machine at $t$ \footnote{When multiple machines are available at $t^{th}$ transition, we randomly sample the machine such that a transition always loads an operation to a machine and we increment transition index $t$} or (2) skipping scheduling at $t$ because no operations are eligible to be processed (null action $a_{\emptyset}$). The transition function $\mathrm{P}$ models the transition from the current state $g_t$ to the next state $g_{t+1}$ due to the action $a_t$. The reward function $\mathrm{R}$ models the immediate reward given by a simulator from every transition. $\gamma$ is the discount factor. The initial state is the snapshot of the job shop where none of the jobs are processed. The terminal state is the snapshot when all jobs are processed. 

\subsection{Efficient MDP formulation of JSSP}
While the MDP formulation of the JSSP suggests a framework to consider the time-dependent nature of the JSSP, consideration of all transitions is not computationally attractive. We observed that a significant number of the unit-time transitions are driven by $a_{\emptyset}$. Such transitions are trivial when we assume the transition model is deterministic. For instance, when all machines are processing the operations, the transitions until at least one machine becomes available will be driven by $a_{\emptyset}$ while a unit-time transition continuously is accruing. From this observation, we propose a new MDP setup which models the transitions only due to the actions $a_t \neq a_{\emptyset}$. We first define the non-trivial states as the states when (1) at least one machine is available and (2) a machine has strictly more than one operation ready to be processed. For brevity, we will refer to the original transitions as primitive transitions and non-trivial states as states. According to the new state definition, we propose the following transition model and corresponding reward function:
\begin{equation}
\label{eqn:event_based_transtion}
P(g,a,g') \triangleq \mathbb{P}(g_{\tau+1}=g' \vert g_{\tau}=g, a_{\tau}=a)
\end{equation}

\begin{align}
\label{eqn:event_based_reward}
R(g_{\tau},a_{\tau},g_{\tau+1}) &= \sum_{t=0}^{t=t(\tau+1)-t(\tau)} \gamma^{t} R_t(g_t, a_t, g_{t+1}) \\
&= \sum_{t=0}^{t=t(\tau+1)-t(\tau)} \gamma^{t} R_t(g_t, a_{\emptyset}, g_{t+1})
\end{align}
Here, $\tau$ denotes the index of the non-trivial state transition from the initial state. $t(\tau)$ is the index of primitive transition when the $\tau^{\mathrm{th}}$ non-trivial transition happens.

We ignore the primitive transitions and corresponding state-action pairs between two consecutive non-trivial states because the actions for the primitive transitions are null (note that the case $a_t = a_{\emptyset}$ is not subject to training). The second equality on \eqref{eqn:event_based_reward} holds because only null actions are executed between two non-trivial transitions.

The MDP formulation with non-trivial transitions can be seen as a special case of semi-MDP formulation \cite{sutton1999between} with the primitive transitions occuring until the next state is non-trivial. Due to the semi-MDP formulation with the transition model \eqref{eqn:event_based_transtion} and the reward function \eqref{eqn:event_based_reward}, the RL framework can solve the original MDP with the presence of non-trainable actions $a_{\emptyset}$.

\begin{figure}[t]
  \centering
  \includegraphics[width=0.75\textwidth]{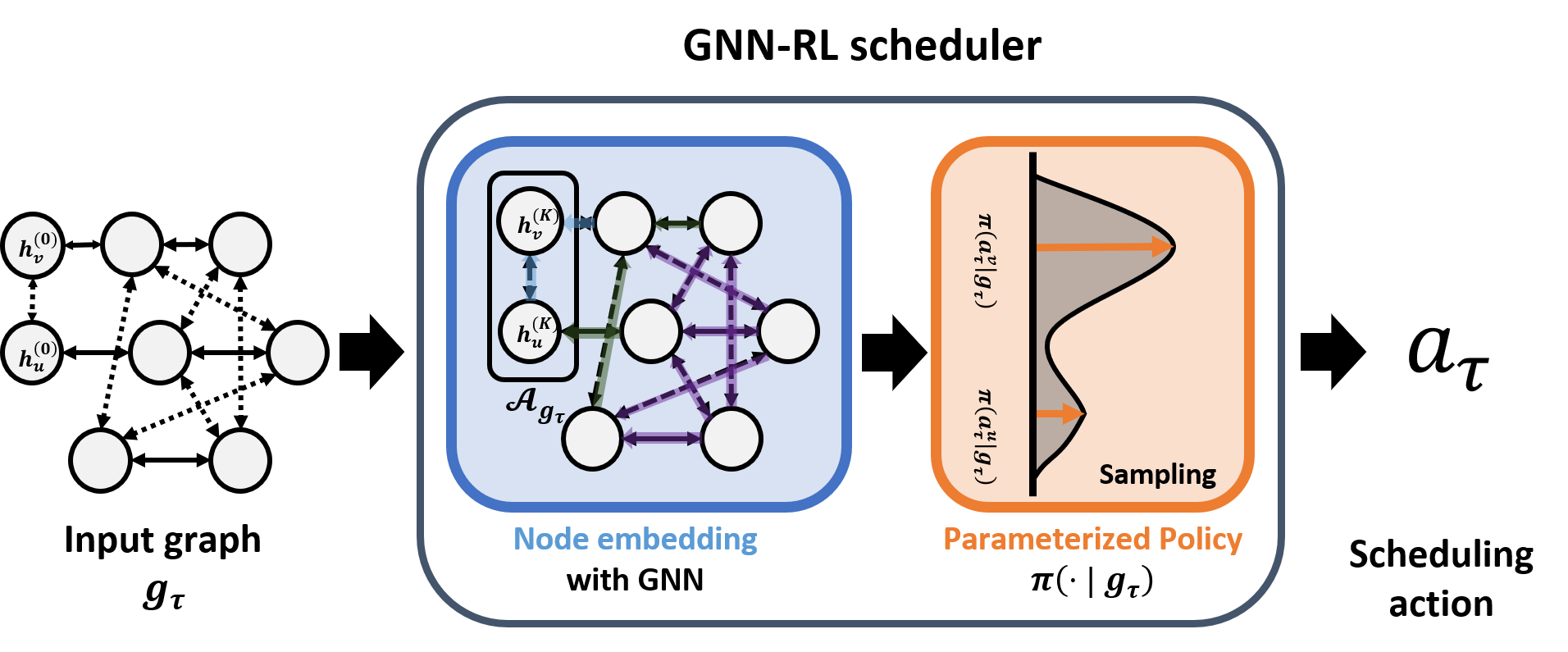}
  \caption{\textbf{Overall framework for computing the scheduling action}}
  \label{fig:methodology_overview}
\end{figure}

\section{Methodology}
In this section, we describe the overall framework including how a scheduling action is derived from the given JSSP state $g_\tau$. The overall framework is composed of (1) a representation module that generates the node embedding from the input graph $g_t$ and (2) a dispatching module that takes the embedding and computes the probability distribution of the feasible actions for the nodes in $g_\tau$. The entire procedure for the computing scheduling action is illustrated in Figure~\ref{fig:methodology_overview}

\subsection{Node embedding procedure}

Graph node embedding is a method of representing nodes in continuous vector space, considering node/edge feature and preserving different types of relational information form the graph. The node embedding can then be thought of as a feature vector containing sufficient information about the target node and its relational information in the graph, which can be readily used to conduct various end-tasks. We compute the node embedding by iteratively applying an embedding layer.

We propose a embedding layer that is designed for explicitly consider the different relations among the nodes in the JSSP graphs. The proposed embedding layer maintains four differentiable functions: precedent node updater $f_p(\cdot \ ; \theta_1)$, succedent node updater $f_s(\cdot \ ; \theta_2)$, disjunctive node updater $f_d(\cdot \ ; \theta_3)$, and target node updater $f_n(\cdot \ ; \theta_4)$. The computation procedure of the proposed node embedding layer is given as follows:

\begin{equation}
\label{eqn:gnn_computation}
\begin{aligned}
h_{v}^{(k)} & = f_{n}^{(k)}
\left( 
\mathrm{ReLU}\left(f_p^{(k)}\left(\sum_{i \in \mathcal{N}_{p}(v)} h_{i}^{(k-1)}\right)\right) \bigg\|
\text{ } \mathrm{ReLU}\left(f_s^{(k)}\left(\sum_{i \in \mathcal{N}_{s}(v)} h_{i}^{(k-1)}\right)\right) \bigg\| \right. \\
&\left. \hspace{.5in} \text{ } \mathrm{ReLU}\left(f_d^{(k)}\left(\sum_{i \in \mathcal{N}_{d}(v)} h_{i}^{(k-1)}\right)\right) \bigg\|
\text{ }\mathrm{ReLU}\left(\sum_{i \in \mathcal{V}} h_i^{(k-1)}\right) \bigg\|
\text{ }h_v^{(k-1)} \bigg\|
\text{ }h_v^{(0)} 
\right)
\end{aligned}
\end{equation}
where $\mathrm{ReLU}(x) = max(0,x)$ position-wisely, and $\|$ is the vector concatenation operator. $\mathcal{N}_p(v)$ and $\mathcal{N}_s(v)$ denote the precedent and succedent node set of node $v$, respectively. $\mathcal{N}_d(v)$ is the disjunctive neighborhood of node $v$. i.e., the set of nodes connected to $v$ by disjunctive edges. $h_v^{(k)}$ is the $k^{th}$ updated embedding of node $v$. $h_v^{(0)}$ is the initial node feature $x_v$ by definition. \newline

The proposed embedding layer generates the updated node embedding $h_v^{(k)}$ while utilizing six different types of inputs that are for accounting various relationships among the nodes. The inputs are as follows:
\begin{itemize}
    \item The precedent node embedding
    $\sum_{i \in \mathcal{N}_{p}(v)} h_{i}^{(k-1)}$. The precedent node embedding is used as the input of the precedent node updater $f_p(\cdot \ ; \theta_1)$. The precedent node updater $f_p(\cdot \ ; \theta_1)$ generates an intermediate embedding that contextualizes the relationship between target node $v$ and the precedent nodes in $\mathcal{N}_{p}(v)$.
    
    \item The succeeding node embedding
    $\sum_{i \in \mathcal{N}_{s}(v)} h_{i}^{(k-1)}$. The succeeding node embedding is used as the input of the succedent node updater $f_s(\cdot \ ; \theta_2)$ to generate intermediate node embedding capturing the relationships between target node $v$ and the succeeding nodes in $\mathcal{N}_{s}(v)$.
    
    \item The disjunctive node embedding
    $\sum_{i \in \mathcal{N}_{d}(v)} h_{i}^{(k-1)}$. The disjunctive node embedding is used as the input of the disjunctive node updater $f_d(\cdot \ ; \theta_3)$ to generate intermediate node embedding capturing the relationships between target node $v$ and the disjunctive nodes in $\mathcal{N}_{d}(v)$.
    
    \item The graph-level embedding
    $\sum_{i \in \mathcal{V}} h_i^{(k-1)}$. The graph-level embedding is used to consider the entire status of a job shop.
    
    \item The previous embedding $h_v^{(k-1)}$. The previous embedding is used to differentiate the specific node $v$ from other nodes in the graph.
    
    \item The initial node embedding $h_v^{(0)}$. The initial node feature is used to consider the initial input feature of the target node.
\end{itemize}
Using the previous embedding $h_v^{(k-1)}$ and all the intermediate node embeddings described above, the target node updater function $f_n(\cdot \ ; \theta_4)$ computes the updated node embedding $h_v^{(k)}$.

We set the node embeddings $h_i^{(k)}$ as zero vectors with the same dimensions as $h_v^{(k)}$ when the corresponding operation have finished or is non-exist. Such modification enables the proposed embedding layer to compute the node embedding with any JSSP graphs. Apart from the computational aspect, the combination of zero-vector modification and $\mathrm{ReLU}(\cdot)$ encourages the feature updaters $f_p(\cdot \ ; \theta_1)$, $f_s(\cdot \ ; \theta_2)$, and $f_d(\cdot \ ; \theta_3)$ to learn sparse representations of node embedding features \cite{yang2018glomo}. That is, when the input node features do not significantly impact the scheduling performance, the embedding layer presumably processes input node features in the same manner as non-existing nodes.

\begin{figure}[t]
  \centering
  \includegraphics[width=0.8\textwidth]{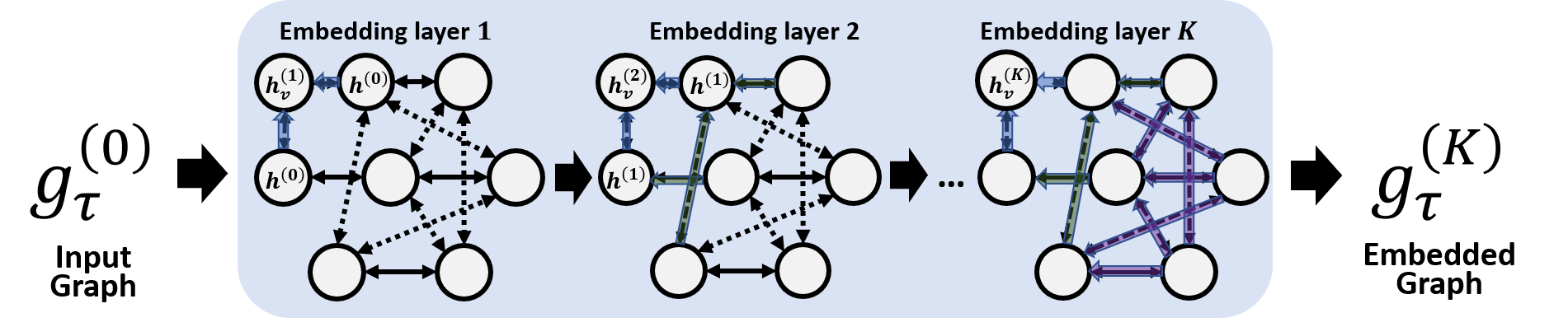}
  \caption{\textbf{GNN with $K$ embedding layers} White circles denote the operations. Straight and dotted edges indicate the conjunctive edges and the disjunctive edges, respectively. 
  The source nodes of blue, green, and purple colored shadows are 1, 2, and 3 hop neighbors of the target node $v$. (Best shows in color)}
  \label{fig:GNN}
\end{figure}

We construct a GNN by stacking $K$ embedding layers. Through $K$ embedding iterations, we can compute the embedded graph $g^{(K)}_{\tau}$ from the input graph $g^{(0)}_{\tau}$. The node features of $g^{(K)}_{\tau}$ contain information of their $K$-hop neighborhood as shown in Figure~\ref{fig:GNN}. The graph-level embedding of $g^{(K)}_{\tau}$ provides representation of the graph such that the graph is distinguishable from other disjunctive graphs of JSSP instances \cite{xu2018powerful}. For instance, the magnitude of $\sum_{i \in \mathcal{V}} h_i^{(K)}$ can provide a clue to the size (number of nodes) of input graph $g_\tau$.

\subsection{Decision making with node embedding}
The node embeddings computed by the proposed GNN are used to choose the optimal scheduling action. We assume that the node embeddings is effective for deciding the optimal scheduling actions. We train the GNN parameters to ensure that the computed node embedding produces good scheduling results. The detailed training procedure is discussed in the next section. 

In the MDP formulation for JSSP, ‘action $a_{\tau}$’ is assigning one of the available jobs to the target machine. It is equivalent to choose one of the nodes (operations) that can be processed by the target machine. The state $g_{\tau}$ contains information on which machine is available at the $\tau^{th}$ transition. 

To derive the scheduling policy, we introduce actor $\pi\left(a_{\tau} | g^{(K)}_{\tau}\right)$. The actor produces the probability distribution over the available (feasible) operations to be chosen by the target machine using the softmax function as follows:
\begin{equation}
\label{eqn:actor}
\pi(a_\tau^v | g^{(K)}_{\tau}) \triangleq \frac{exp(f_l(h_v^{(K)};\theta_5))}{\sum_{u \in A_{g_\tau}} exp(f_l(h_u^{(K)};\theta_5))}
\end{equation}
where $f_l(\cdot \ ; \theta_5)$ is a differentiable function that maps node embeddings to the logit values of each node and  $\mathcal{A}_{g_\tau}=\left\{a_{\tau}^{v} \ ; v \in \mathrm{processible\ nodes\ at}\ \tau \right\}$ is the processible action set \footnote{We overload the notation of action $a_{\tau}$ as $a_{\tau}^{v}$ to clearly show that the action is ‘selecting the node $v$’.}. One of the feasible scheduling actions is then selected from this stochastic policy. 

From our experience, we confirmed that JSSPs have multiple (sub) optimal scheduling trajectories. The proposed stochastic policy allows the actor to explore multiple (sub) optimal scheduling trajectories and chose one of them.

To train the scheduling policy $\pi(a_\tau^v | g^{(K)}_{\tau})$, we employ proximal policy optimization (PPO) \cite{schulman2017proximal}, a policy gradient-based reinforcement learning algorithm. Since the PPO algorithm uses a critic that approximates the state value of an MDP, here we present the form of the critic function (the detailed procedure how the parameters of the policy and the critic along with the parameters of GNN will be discussed in the next section). We approximate the state-value by using critic $V\left(g^{(K)}_{\tau}\right)$ as follows:
\begin{align}
\label{eqn:critic}
V^\pi\left(g^{(K)}_{\tau}\right) \approx V\left(g^{(K)}_{\tau}\right) = f_{v}\left(\sum_{i \in \mathcal{V}} h_i^{(K)};\theta_6\right)    
\end{align}
where $\sum_{i \in \mathcal{V}} h_i^{(K)}$ is graph-level embedding, and $f_{v}(\cdot \ ; \theta_6)$ is a differentiable function.

\section{Training procedure}
\label{section:training}
In this section, we describe the training procedure of the proposed GNN and decision making modules. The parameters of model $\Theta = \{\theta_1, \theta_2,\theta_3,\theta_4,\theta_5,\theta_6\}$ are optimized using RL framework. The RL framework requires the sample of transitions $(g_\tau, a_\tau, r_\tau, g_{\tau+1})$ to train the model parameters. We collect the samples of transitions by simulating job shop using our own developed JSSP simulator.

\subsection{\textbf{pyjssp}: MDP formulated JSSP simulator}
The simulator handles not only the state transition induced by the scheduling action, but also transformation from the state of the JSSP instance to the corresponding disjunctive graph. Moreover, according to the proposed MDP formulation, the simulator automatically transits the trivial actions until a non-trivial event occurs and computes the cumulative reward, ${R(g_{\tau},a_{\tau},g_{\tau+1})}$. For simulating this particular JSSP, we implemented a simulator, \textbf{pyjssp}\footnote{Refer the webpage
\href{https://pypi.org/project/pyjssp}{https://pypi.org/project/pyjssp}}. The simulator emits a transition sample ${(g_{\tau},a_{\tau},R_{\tau},g_{\tau+1})}$ for every non-trivial scheduling action.

\subsection{Design of reward function}
We aim to learn a scheduling policy that minimizes the makespan $C_{max}$. $C_{max}$ is the maximum completion time among jobs, i.e., the total length of the schedule ($C_{max}=\max_{j\in J}C_j$). To train the scheduling policy to minimize $C_{max}$, we employ the waiting-job reward function as follows:
\begin{align}
\label{eqn:waiting_time_reward}
r(g_{t},a_{t},g_{t+1}) = -(\text{the number of waiting jobs at time } t)
\end{align}
The waiting-job reward design is adequate for achieving a minimal makespan. Maximizing the cumulative sum of rewards, i.e., minimizing the cumulative sum of the number of waiting jobs, is strongly associated with minimizing the makespan \cite{pinedo2008scheduling}.

The suggested reward representation is beneficial for training the proposed model with the RL algorithm. The waiting-job rewards are well-defined in every transitions; thus, the RL algorithm will frequently obtain meaningful reward signal during training. On the other hand, the makespan reward will be realized once only when the given instance is finished; as a result, the RL algorithm has difficulty in assigning credit to the actions in the episode during training. 

\subsection{Reinforcement learning algorithm}
We employ proximal policy optimization (PPO) \cite{schulman2017proximal}, which is a promising variant of the policy gradient method, to train the proposed model. Policy gradient methods optimize the policy by updating parameter set $\Theta$ in the direction of the gradient of the objective function $\nabla L(\Theta)$:
\begin{align}
\Theta_{new} = \Theta_{old}+\eta\nabla_{\Theta}L(\Theta)
\end{align}
where $L(\Theta)=\hat{\mathbb{E}}_{s\sim{d^{\pi}}}\left[\sum_{a\in A}\pi_{\Theta}(a|s) A^{\pi}(s,a)\right]$, and $\eta$ is the learning rate.

When updating parameters with the original policy gradient method, small changes in the parameter space often lead to substantial changes in the policy space, which results in unstable training. In particular, during job shop scheduling, this property is critical because the state of the JSSP cannot relapse into the previous state, and a single action significantly affects the performance of the schedule. Therefore, to overcome this drawback of policy gradient methods, trust region policy optimization (TRPO) \cite{schulman2015trust} has been proposed to limit the size of policy update between $\Theta_{old}$ and $\Theta_{new}$ using a KL divergence constraint. However, employing TRPO requires solving a constrained optimization to update the parameters, which is computationally demanding. As an alternative approach, PPO simplifies this concept by using a clipped surrogate objective. PPO updates the parameters of the GNN and the policy only when the current representation and the policy module (locally) improve the scheduling performance. Specifically, the clipped surrogate objective is defined as:
\begin{align}
\label{eqn:ppo_loss}
L_{\tau}^{\text{CLIP}}(\Theta)=\hat{\mathbb{E}}_{\tau}
\left[\min 
    \left
        (r_{\tau}(\Theta) \hat{A}_{\tau}, \text{clip}(r_{\tau}(\Theta), 1-\epsilon, 1+\epsilon) \hat{A}_{\tau}
    \right)
\right]
\end{align}
where $r_{\tau}(\Theta)=\frac{\pi\left(\mathbf{a}_{\tau} | g_{\tau} ; \Theta_{\text {new }}\right)}{\pi\left(\mathbf{a}_{\tau} | g_{\tau} ; \Theta_{\text {old }}\right)}$ and $\Theta = \{\theta_1, \theta_2,\theta_3,\theta_4,\theta_5,\theta_6\}$.

The function $\operatorname{clip}\left(r_{\tau}(\Theta), 1-\epsilon, 1+\epsilon\right)$ limits $r_{\tau}(\Theta)$ within $[1-\epsilon, 1+\epsilon]$, and $\hat{A}_{\tau}$ is the generalized advantage estimation function defined as follows:
\begin{align}
\label{eqn:gen_adv_func}
\hat{A}_{\tau}=\delta_{\tau}+(\gamma \lambda) \delta_{\tau+1}+\cdots+(\gamma \lambda)^{T-\tau-1} \delta_{T-1}
\end{align}
where $\delta_{\tau}=r_{\tau}+\gamma V\left(g_{\tau+1} ; \Theta\right)-V\left(g_{\tau} ; \Theta\right)$, and $T$ is the termination step of the episode.

In addition, we combine a value function error and entropy bonus term with the objective function to enhance the training performance. 
The advantage function can indicate the correct direction in which to update the parameters when the estimation of the value function is accurate. The entropy bonus encourages sufficient exploration during training. Thus, the combined objective function is as follows:
\begin{align}
\label{eqn:ppo_loss_total}
L_{\tau}^{total}(\Theta)=\widehat{\mathbb{E}}_{\tau}\left[L_{\tau}^{\text{CLIP}}(\Theta)-\alpha\left(V\left(g_{\tau} ; \Theta\right)-V_{\tau}^{t a r g}\right)^{2}+\beta S_{\tau}\left(\pi_{\Theta}\right)\right]
\end{align}
where $S_{\tau}(\pi_\Theta) = -\sum_{a \in \mathcal{A}_{g_\tau}} \log\left(\pi_\Theta(a)\right) \pi_\Theta(a)$ denotes the entropy of current policy $\pi_{\Theta}$ at $\tau$; $V_{\tau}^{\mathrm{targ}}=\sum_{i=\tau}^{T} r_i$ is the realized sum of rewards; and $\alpha$ and $\beta$ are coefficients. 

The PPO algorithm maximizes \eqref{eqn:ppo_loss_total} by updating the set of parameters $\Theta$ in the direction suggested by $\nabla_\Theta L_{\tau}^{total}(\Theta)$. By repeatedly updating the set of parameters, the actor policy $\pi_{\Theta}(a_{\tau}|g_{\tau})$ is trained to maximize the expected cumulative sum of rewards $V(g_{\tau};\Theta)$.

\subsection{Training details}
Here, we detail the hyper-parameters used during training. Across the models, we use multi-layer perceptrons (MLP) with two hidden layers of 256 $\mathrm{ReLU}$\cite{glorot2011deep} activation units and appropriate input and output dimensions to compute embeddings, logits of nodes, or value of graphs. 
\footnote{ReLU network is a standard building block for complex neural networks. Choosing $2^n$ numbered hidden layers is also standard practice in the deep learning community. Similarly, we set the dimension of hidden node embedding as $2^3$. We tested deeper and thicker ReLU networks while finding the best hyperparameters. The deep and thick ReLU networks than the specified hyperparameter setup did not improve scheduling performance.}
In the GNN, the precedent, succedent, and disjunctive node updaters $f_p^{(k)}(\cdot \ ; \theta_1)$, $f_s^{(k)}(\cdot \ ; \theta_2)$, and $f_d^{(k)}(\cdot \ ; \theta_3)$ have 8-dimensional inputs and outputs. As a result, the node updater $f_n^{(k)}(\cdot \ ; \theta_4)$ has 48-dimensional input and produces 8-dimensional node embeddings. 

The decision making modules (actor and critic) also employ MLPs to compute the logits and values of the given JSSP graph. $f_l(\cdot \ ; \theta_5)$ and $f_{v}(\cdot \ ; \theta_6)$ take 8-dimensional inputs and return scalar values. We set the number of embedding iterations as $K=3$. $K$ controls the scope of information propagation when computing node embeddings; the node embedding of the target node is computed using the other node embeddings of $K$-hop neighborhood nodes. When larger $K$ is used, the GNN module can compute the target node embedding while considering more high-order interactions among the nodes. However, we experimentally confirmed that $K$ larger than three does not significantly improve the GNN scheduler's scheduling performance but only requires more computations. We verified $K=3$ was enough for even our largest testing JSSP, which has 20 machines and 100 jobs, because disjunctive graphs are densely connected.

We randomly generate the initial graph $g_0$ and performs scheduling to collect transition samples $(g_\tau, a_\tau, r_\tau, g_{\tau+1})$. The initial state distribution $\mathbb{P}_0(\mathcal{G}_0)$ randomly generates $g_0$ by specifying the number of machines $m$ drawn from a uniform integer distribution between 5 and 9, i.e., $m \sim \mathcal{U}(5, 9)$, and the number of jobs $n \sim \mathcal{U}(m, 9)$ and sample processing times from $\mathcal{U}(1, 99)$. We randomly permute the order of machines and assign a machine to operations of a job such that operations of a job are processed by different machines. 

\begin{table}[]
\centering
\caption{\textbf{PPO hyperparameter used for training the GNN scheduler}}
\vskip 0.09in
\begin{tabular}{l|c}
\toprule
\textbf{Hyperparameters}              & \textbf{Value}    \\ 
\midrule
Optimizer                             & Adam\cite{kingma2014adam} \\
Learning Rate ($\eta$)                & $2.5\times 10^{-4}$ \\
Discount factor ($\gamma$)            & 1.0 \\
GAE parameter ($\lambda$)             & 0.95 \\
Clipping parameter ($\epsilon$)       & 0.2 \\
value function coefficient ($\alpha$) & 0.5 \\
Entropy bonus coefficient ($\beta$)   & 0.01 \\
Number of episodes per update ($n$)   & 20 \\
\bottomrule
\end{tabular}
\label{table:RL_hyperparameters}
\end{table}

We collect the transition samples from 20 episodes with the 20 JSSP instances using the current scheduling policy. After sample collection, we update $\Theta$ with the PPO algorithm. For every five updates, we regenerate another random 20 JSSP instance. We terminate the training procedure when the makespan of the validation set converges. We prepare the validation instances generated through $p_0(\mathcal{G}_0)$. The optimal solutions (minimum makespan) of the validation instances are computed through the Google OR tool~\cite{google-or-tools}. Algorithm~\ref{algorithm:RL} describes the entire training procedure.

The hyperparameters of the PPO algorithm can affect the performance of the trained scheduling policy. Table~\ref{table:RL_hyperparameters} summarize the hyperparameters of the PPO algorithm, whose values are carefully selected to achieve reliable and good performance in the validation JSSP instances. For the learning rate $\eta$, GAE parameter $\lambda$, clipping parameter $\epsilon$, value function coefficient $\alpha$, and entropy bonus coefficient $\beta$, we used the same values suggested by ~\cite{stable-baselines}. This set of hyperparameter values is widely known for working generally well for various problems.  

We modify two hyperparameters, discount factor $\gamma$, and the number of episodes per gradient updates $n$, explicitly considering the characteristic of the JSSP scheduling problem. We set $\gamma$ as 1.0 since the target JSSP problem is formulated as a finite MDP. Besides, we set the number $n$ of episodes per parameter update as 20 such that the PPO algorithm reliably estimates the state value and compute the gradient from it. Due to the combinatorial nature of the JSSP, single action in the early phase of MDP can induce a considerable difference in the result of a remaining MDP. Thus, it is essential to use enough number of the episode when updating the parameter. From a series of experiments, we have observed that when $n$ is less than 20, the computed state value has a large variance, while when $n$ is over 20, the performance gain is not significant. Thus, to achieve the best performance while reducing the computational burden, we used $n=20$.

\begin{algorithm}[h]
\SetAlgoLined
\caption{Training procedure for the GNN scheduler}
\SetAlgoLined
\KwResult{the Trained GNN scheduler}
generate $n$ JSSP instances with the initial graphs $g_0 \sim \mathbb{P}_0(\mathcal{G}_0)$ \\
 \For{iteration=1,2,...}
 {
 \For{episode $i$=1,2,...$n$}{
 \While{the episode is not terminated}{
 \If{the current state is non-trivial state}{
 observe $R_{\tau-1}$and $g_{\tau}$
 
 collect transition sample ${(g_{\tau-1},a_{\tau-1},R_{\tau-1},g_{\tau})} $
 
 execution action $a_{\tau} \sim \pi_{\Theta}(\cdot|g_{\tau})$ 
 }
 Compute the generalized advantage estimates $\hat{A}_1,... \hat{A}_T$
 }}
 Compute total objective function, $L^{\mathrm{total}} (\Theta) = \sum_{i=1}^{n}\sum_{\tau =1}^{T}L_{\tau}^{i}(\Theta)$
 
 Update all parameters, $\Theta \leftarrow \Theta + \eta \nabla_\Theta L^{total}(\Theta)$
 
 \If{Every 5 gradient updates}{
 Regenerate the $n$ JSSP instances with the initial graphs $g_0 \sim \mathbb{P}_0(\mathcal{G}_0)$
 }
  \If{validation performance converges}{
 Terminate training procedure
 }
}
\label{algorithm:RL}
\end{algorithm}

\section{Experiments}
\begin{figure}[t]
  \centering
  \includegraphics[width=0.7\textwidth]{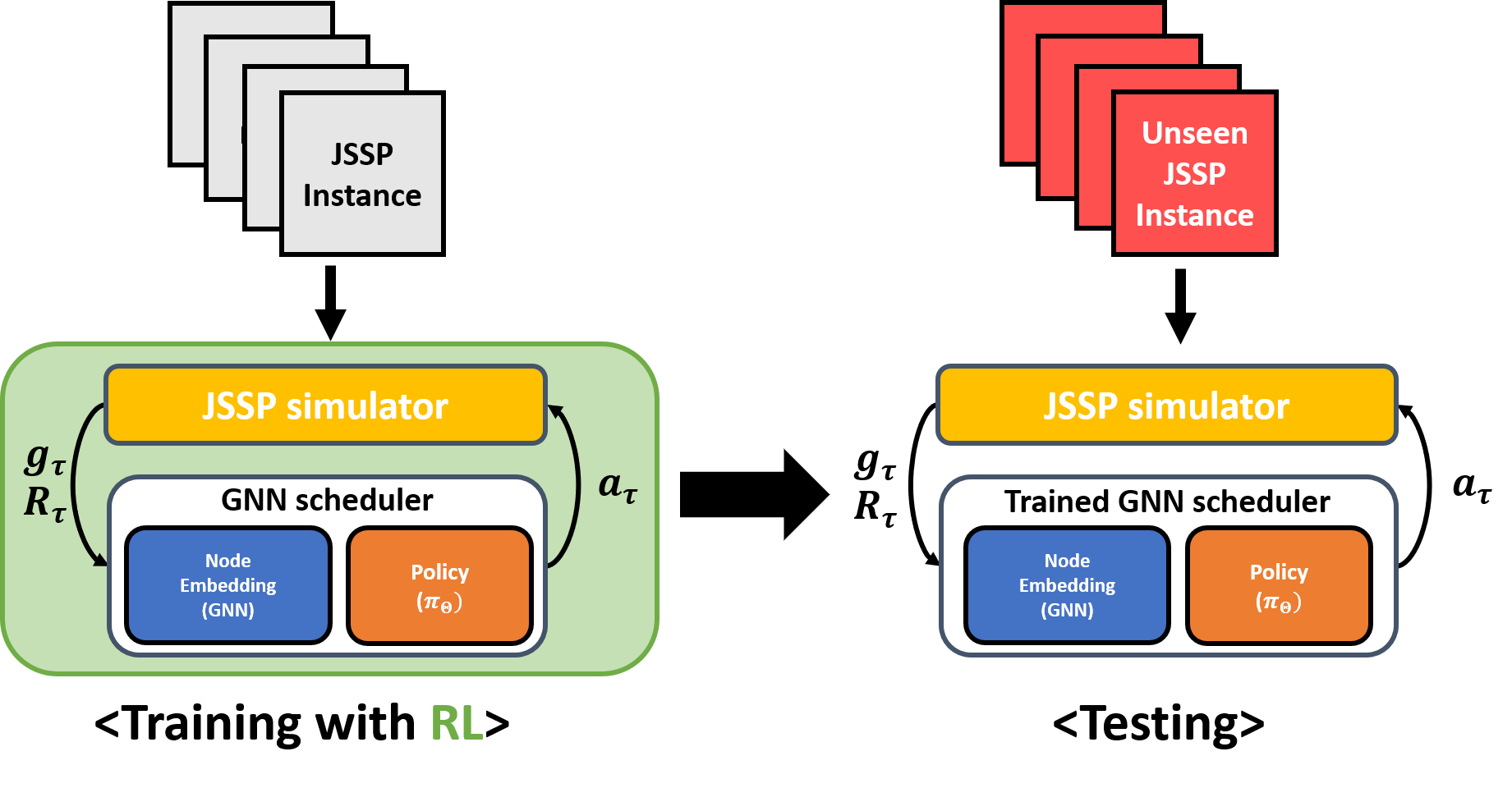}
  \caption{\textbf{Test scheme to measure generalization performance}}
  \label{fig:test_scheme}
\end{figure}

The trained GNN scheduler can schedule any JSSP without instance-by-instance training. To test the generalization performance over JSSP instances, we measured the scheduling performances over different sets of unseen JSSP instances. The overall test scheme is described in Figure ~\ref{fig:test_scheme}. To the best of our knowledge, no learning-based JSSP scheduler is able to schedule on a new JSSP instance that is different from the training JSSP instances in terms of size, constraints, and processing times. Consequently, we compared the generalization performance of our model with the following PDRs: 
\begin{itemize}
  \item First In First Out (FIFO): The first job in is processed first.
  \item Last In First Out (LIFO): The last job in is processed rfirst.
  \item Shortest Processing Time (SPT): The job with the shortest processing time is processed first.
  \item Longest Processing Time (LPT): The job with the longest processing time is processed first.
  \item Shortest Total Processing Time (STPT): The job with the shortest total processing time is processed first.
  \item Longest Total Processing Time (LTPT): The job with the longest total processing time is processed first.
  \item Least Operation Remaining (LOR): The job with the least remaining operations to be completed is processed first.
  \item Most Operation Remaining (MOR): The job with the most remaining operations to be completed is processed first.
  \item Least Queue Next Operation (LQNO): The job whose next operation has the least waiting jobs is processed first.
  \item Most Queue Next Operation (MQNO): The job whose next operation has the most waiting jobs is processed first.
  \item Random: Job priority is randomly assigned.
\end{itemize}

Please note that we have included a set of well-known PDRs as baseline algorithms to compare the performance of the proposed RL-based approach. Although there are sophisticated priority rules such as a search-algorithm optimized CDR \citep{ozturk2019extracting, tay2008evolving}, we have not considered such algorithms because these approaches require applying sophisticated optimizations or search based methods to find the optimum mixing ratio of various PDRs. Finding the best CDR is beyond the scope of the current study. The present study's focus is to empirically investigate the potential of the RL-based approach for deriving an efficient dynamic dispatching rule for JSSP, rather than seeking the algorithm that has the best performance.

The key advantages of the RL-based approach over the PDR and CDR are that it learns how to adaptively utilize various state features in determining the optimal scheduling actions, while PDRs and CDRs constantly use the predetermined same state features in making the next scheduling actions. Dynamically extracting the most relevant features out of the current state and mapping these features to the optimal action can be analogous to using an infinite number of PDRs (because the RL approach does not specify the set of features but adaptively construct) or using CDRs with dynamically varying mixing ratios (because the RL approach considers different aspects of the current state of the JSSP in decision-making). These flexibility and adaptability are the essential factors improving the performance. The GNN based state representation module further strengthens the flexibility and adaptability by considering relationships among machines and tasks.

We evaluated the scheduling performance of a scheduler for a JSSP instance by computing the relative scheduling error defined as follows:
\begin{equation}
\epsilon = \left(\frac{\hat C_{max}}{C^{*}_{max}} -1 \right) \times 100 (\%)
\end{equation}
where $\hat C_{max}$ is the makespan derived from the scheduling solution and $C^{*}_{max}$ is the optimum makespan. When a scheduler achieves an optimal schedule, the relative scheduling error $\epsilon$ becomes $0\%$.

We conducted three experiments. The experiments were done with the GNN scheduler that is pre-trained as described in section~\ref{section:training} without additional training, i.e., zero-shot setting, as illustrated in Figure~\ref{fig:test_scheme}. First, we evaluated how well the GNN scheduler was trained using random JSSP instances on a test dataset composed of JSSP instances sampled by the same distribution used to generate training JSSP instances. Secondly, we evaluated how well the GNN scheduler trained in the first experiment performed on various benchmark JSSP instances reported in the literature. Lastly, we investigated the relative competence of the proposed GNN scheduler compared to other RL-based schedulers by employing our method on test data sets that the RL-based schedulers were evaluated on. For all three experiments, we used various PDRs to compare the performance of the proposed method.

\subsection{Performance evaluation on testing distribution}
\label{subsection:exp1}
We measured the performances of the trained GNN scheduler and PDRs over the 100 testing instances sampled from $\mathbb{P}_0(\mathcal{G}_0)$, which were used to generate training instances. The testing instances are accessible from the \textbf{pyjssp} package with the optimal solutions.

\begin{figure}[htbp]
  \centering
  \includegraphics[width=1.0\textwidth]{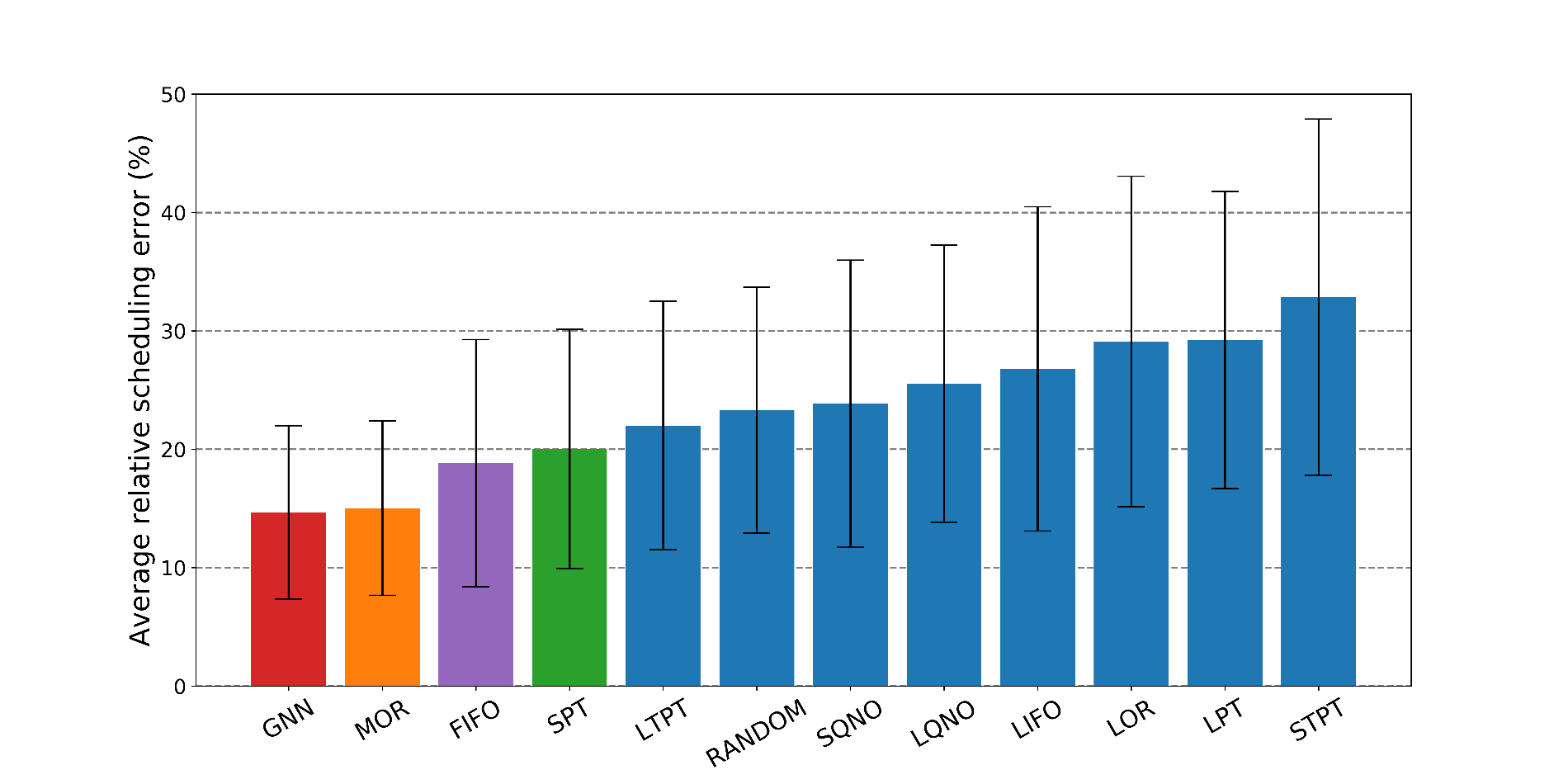}
  \caption{\textbf{Average relative scheduling errors on the testing set} Bar charts describe the average relative error of the GNN scheduler and PDRs. The vertical error bars indicate one standard deviation of errors (the statistics are computed from 100 simulations).}
  \label{fig:exp_1}
\end{figure}

The proposed scheduler outperformed all PDRs on the test instances, as shown in Figure~\ref{fig:exp_1}. The trained GNN scheduler achieved 14.7\% average relative error on the test instances. Among the several PDRs we tested, only MOR achieved a comparable performance with the proposed GNN. The relative scheduling errors of the other PDRs were all over 20\%. This result indicates that the proposed scheduler provides better schedules than all PDRs when solving JSSPs generated by the distribution used to generate JSSP instances for training.

In a practical sense, the distribution $\mathbb{P}_0(g_0)$ used to generate the training and test JSSP instances is not general enough to cover real-world JSSPs. In the following sections, we show that the GNN scheduler trained on this JSSP distribution performs well on JSSPs whose sizes are much larger than the training instances and whose distributions of processing time and machine order are different.

\subsection{Performance evaluation on Benchmark Problems}
\label{subsection:exp2}
In this section, we analyze the performances of the GNN scheduler, trained using the training set described in the previous section, on the following benchmark job shop problems: ORB 01-10 \cite{applegate1991computational}, SWV 01-20 \cite{storer1992new}, FT 06/10/20 \cite{fisher1963probabilistic}, LA 01-40 \cite{lawrence1984resouce}, ABZ5-9 \cite{adams1988shifting}, YN 1-4 \cite{yamada1992genetic}, and TA 01-80 \cite{taillard1993benchmarks}. The numbers of machines and jobs of a JSSP instance in the benchmark sets range from 5 to 20 and 6 to 100, respectively. Each set of benchmark instances follows a different distribution for generating JSSP instances. Refer appendix~\ref{appendix:benchmark_jssp_gen_dists} to find detail description about the generating distribution of benchmarks.

Using the benchmark instances, we investigated how well the trained GNN scheduler solved completely new JSSP instances whose sizes and other characteristics were completely different to those of the training JSSP instances. For some large-size JSSP benchmarks, there are no exact solutions found in the literature \cite{van2018current}. Instead, the lower bound and upper bound of the makespan are only known. For such cases, we consider the mean value of the lower bound and upper bound of the JSSP instance as $C^{*}_{max}$.

\begin{figure}[htbp]
  \centering
  \includegraphics[width=1.0\textwidth]{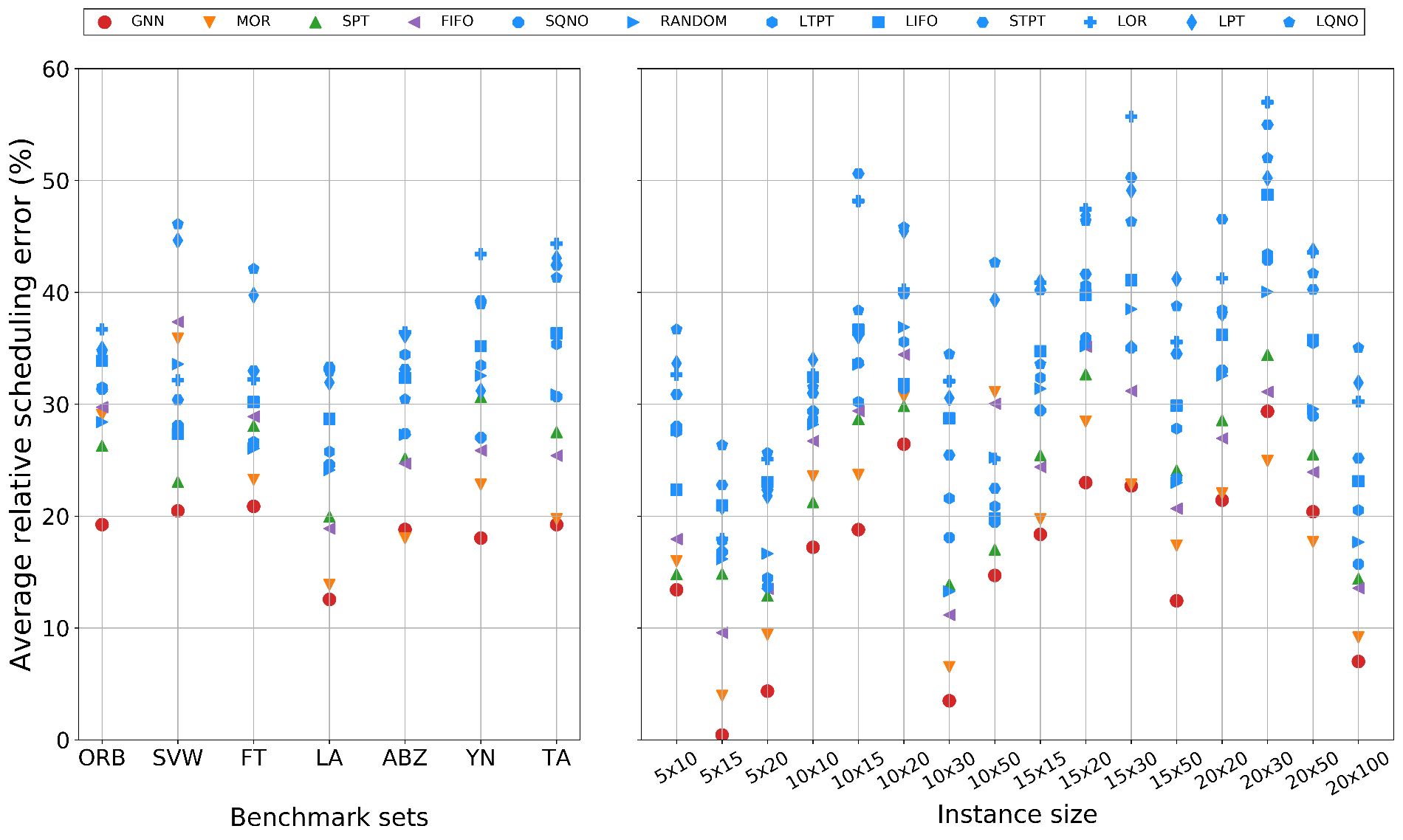}
  \caption{\textbf{Average relative scheduling error on the benchmarks} [Left] Comparison for benchmark type. [Right] Comparison for instance size. Red balls indicate the average relative scheduling errors of the GNN scheduler. Orange, green, and purple marks indicate the average errors of leading PDRs: MOR, SPT, and FIFO, respectively (Best shows in color).}
  \label{fig:exp_2}
\end{figure}

Figure~\ref{fig:exp_2} [Left] shows the average relative error of the GNN scheduler on each benchmark set compared to the PDRs. The GNN scheduler yielded quality solutions whose average relative errors ranged from 12.6\% to 20.9\%. Except the ABZ and TA benchmark JSSPs, the relative errors of the trained GNN scheduler were the lowest. In the cases of ABZ and TA, the difference in relative error between our scheduling policy and the best PDR was within 0.1\%. 

The results of the trained GNN scheduler with respect to the size of JSSP are presented in Figure~\ref{fig:exp_2} [Right]. We excluded the 6$\times$6 (6 machines and 6 jobs) instances from the performance evaluation because the benchmark JSSP with that size is only FT06. Figure~\ref{fig:exp_2} [Right] shows that the proposed GNN scheduler also obtained  strong performances over all JSSPs across all sizes compared to the PDRs. Our scheduler ranked best in all but two size of instances, where it ranked second for the cases of 20$\times$30 and 20$\times$50, with 4.4\% and 2.7\% gaps in relative error from the best PDR, respectively.

From the experiment on the benchmark instances, we can confirm that the proposed GNN scheduler is relatively robust to variations of problem distribution and size compared to PDRs. PDRs schedule operations according to a pre-determined rule without considering the JSSP structure. Therefore, the quality of scheduling solutions from PDRs may fluctuate depending on the properties of the particular JSSP. For example, the MOR and FIFO rules generally performed well, but their scheduling performances deteriorated significantly on the SVW benchmark problems. In contrast, our GNN scheduler reliably generated the high-quality scheduling solutions  for various types of JSSPs. This result implies that the proposed graph embedding algorithm enables the scheduling algorithm to (1) extract effective features representing the current state of JSSP regardless of its size and processing time and (2) derive a better scheduling action.

Also note that, as shown in Figure 6, the error score of the proposed method varies significantly depending on the benchmark instances. The large fluctuation of the error score is possibly caused by (1) the characteristic of a JSSP problem and (2) the large variability in the difficulty level of the benchmark instances.  Because the optimum solution of a JSSP is represented as a unique sequence of assignments, even a single change in the order of assignments can induce a considerable performance degradation, which is a typical property of a function defined over high-dimensional combinatorial action space. Furthermore, this performance degradation strongly depends on the structure of a JSSP (i.e., the distribution of numbers for jobs and machines, and the distribution of task completion time). This argument can be supported by the fact that the error function patterns over the benchmark instances are similar for all the algorithms, including the proposed one. For instance, the scheduling algorithms generally have a low level of an error on LA instances, while the scheduling algorithms have a high level of an error on SVW and FT instances.

\subsection{Comparison of generalization performance for learning-based schedulers}

There has been some research on developing RL-based schedulers for JSSPs \cite{aydin2000dynamic,gabel2012distributed,gabel2008adaptive,lin2019smart}. However, only few of them, such as multi-agent reinforcement learning (MARL) \cite{gabel2008adaptive} and multi-class DQN (MDQN) \cite{lin2019smart}, have shown that the trained RL-based scheduler can be used to schedule new JSSP instances. The details of these transferability tests found in the literature are summarized in Table~\ref{table:RL_method_specs}. Two MARL schedulers were trained with 5$\times$15 and 10$\times$10 instances and tested on LA06-10 and ORB01-10 instances, respectively \cite{lawrence1984resouce,applegate1991computational}. The MDQN scheduler was trained with 15$\times$20 instances and tested on DMU01-05 and DMU41-45 instances \cite{shirazi2016manuscript}. We compared the scheduling performance of the proposed GNN scheduler with those of the RL-based schedulers. Note that the training and test data sets had the same sizes of machines and jobs. Therefore, only a limited generalization test could be conducted for the case where the conjunctive constraints and processing times are different from the training and test data sets.

\begin{table}[t]
\centering
\caption{\textbf{Generalization test setups for the RL-based schedulers, MARL\cite{gabel2008adaptive} and MDQN \cite{lin2019smart}} Both methods could schedule JSSP  instances whose numbers of machines and jobs are the same as those for the training instances. The performance of the MARL method was evaluated using k-fold cross validation scheme on benchmark instances. Each training fold excluded the test instance for the training set. The training set for MDQN was not precisely reported.}
\vskip 0.09in%
\begin{tabular}{@{}c|ccc@{}}
\toprule
\textbf{RL scheduler} & \textbf{Training set} & \textbf{Test set}  & \textbf{Instance size} \\ \midrule
MARL(5x15)   & LA 06-10     & LA 06-10  & 5 x 15        \\
MARL(10x10)  & ORB 01-10    & ORB 01-10 & 10 x 10       \\
MDQN(15x20)  & N/A          & DMU 41-45 & 15 x 20       \\ \bottomrule
\end{tabular}
\label{table:RL_method_specs}
\end{table}

To mimic the experimental procedures reported in \cite{gabel2008adaptive} and \cite{lin2019smart}, we trained the three GNN schedulers ($\mathrm{GNN}_1$, $\mathrm{GNN}_2$, and $\mathrm{GNN}_3$) with the training JSSP instances with sizes 5$\times$15, 10$\times$10 and 15$\times$20, respectively. We then employed the three trained GNN schedulers on the test data set shown in Table~\ref{table:RL_method_specs}. Furthermore, to investigate the generalization performance of the proposed method, we employed the GNN scheduler that was used for the previous experiment, $\mathrm{GNN}_0$, to test the JSSP instances listed in Table~\ref{table:RL_method_specs}.

\begin{figure}[h]
\centering
\subfloat[][Comparison with MARL\newline approach on LA instances]{
\includegraphics[width=0.31\textwidth]{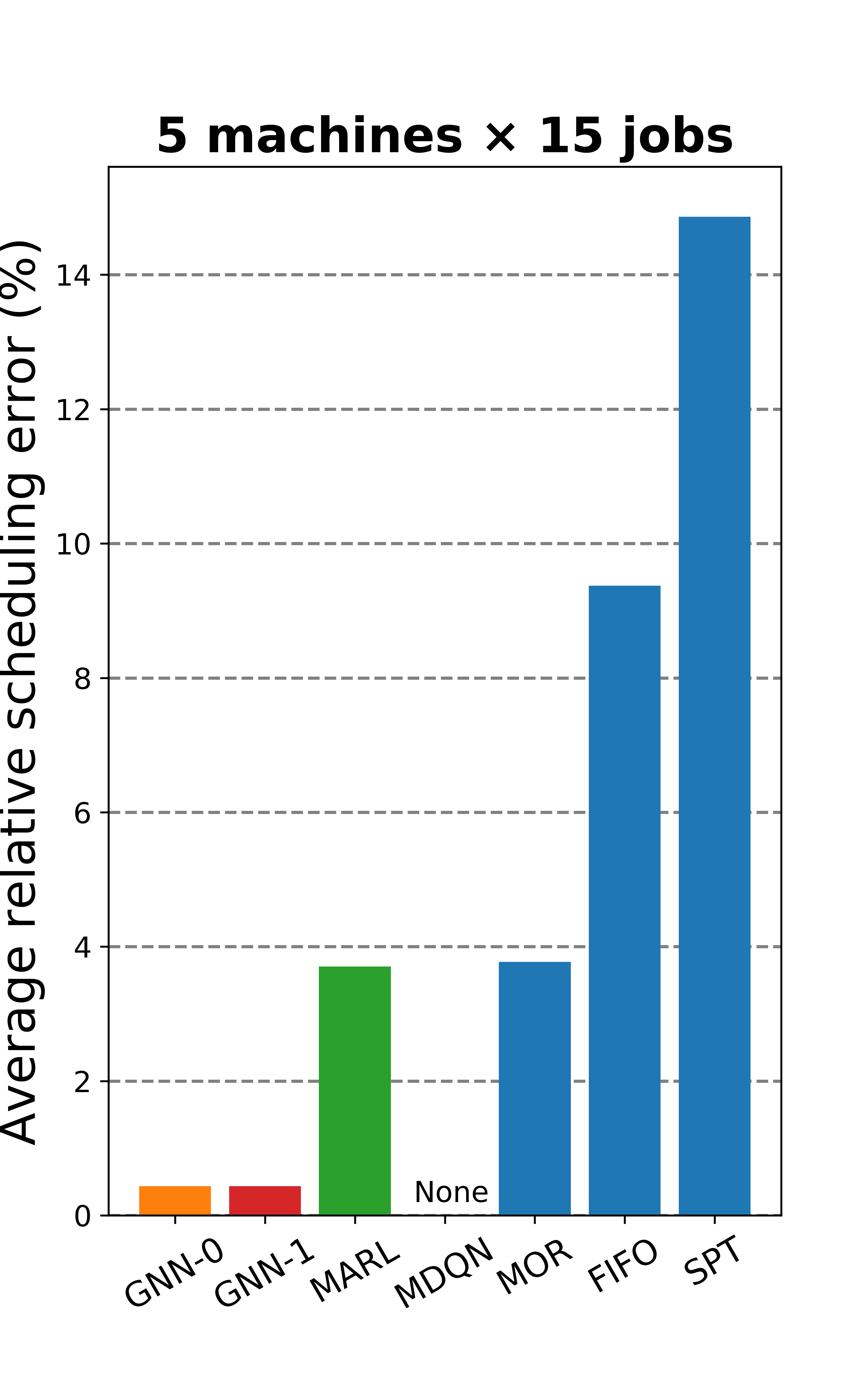}
\label{fig:subfig1}}
\subfloat[][Comparison with MARL\newline approach on ORB instances]{
\includegraphics[width=0.31\textwidth]{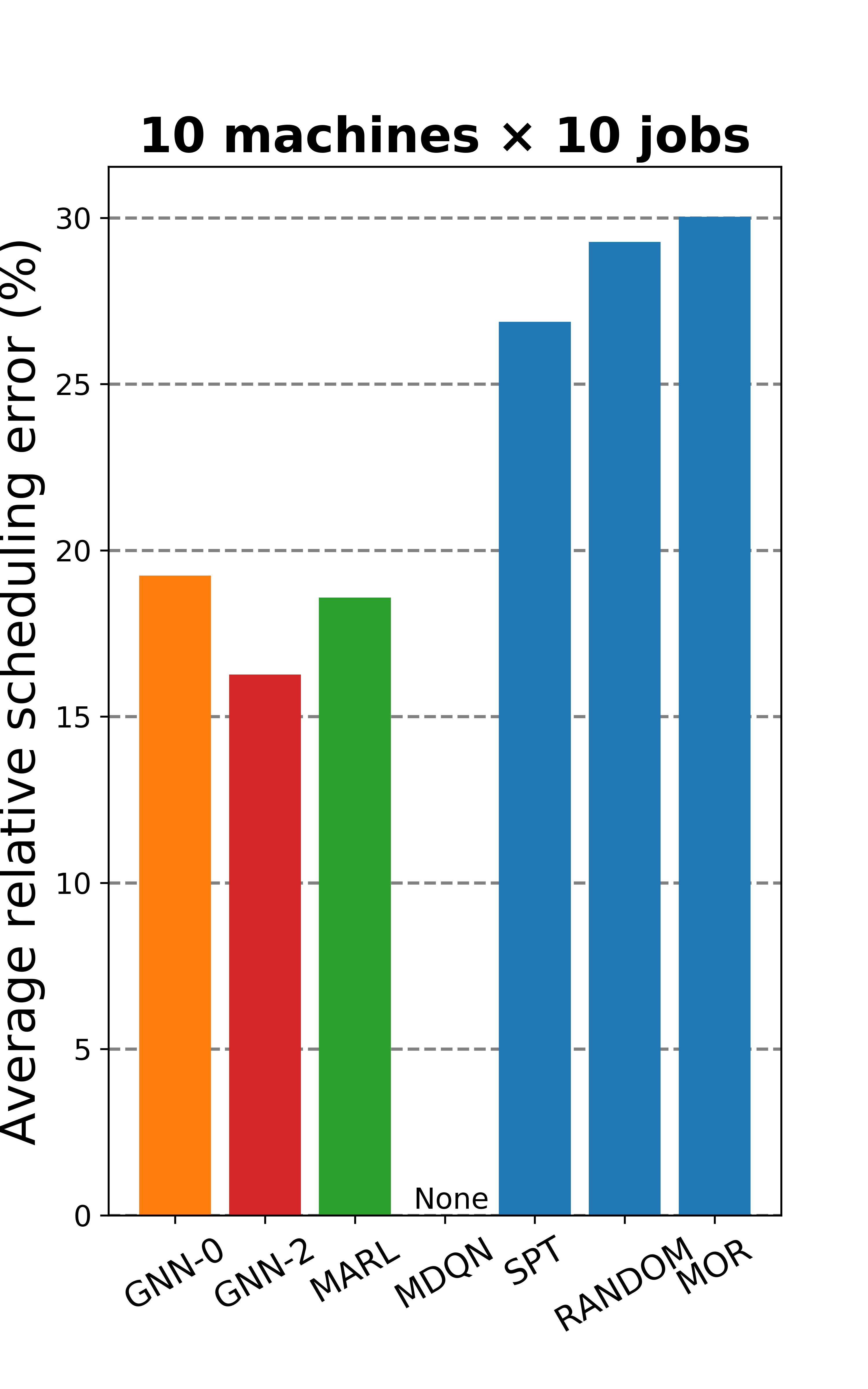}
\label{fig:subfig2}}
\subfloat[][Comparison with MDQN\newline approach on DMU instances]{
\includegraphics[width=0.31\textwidth]{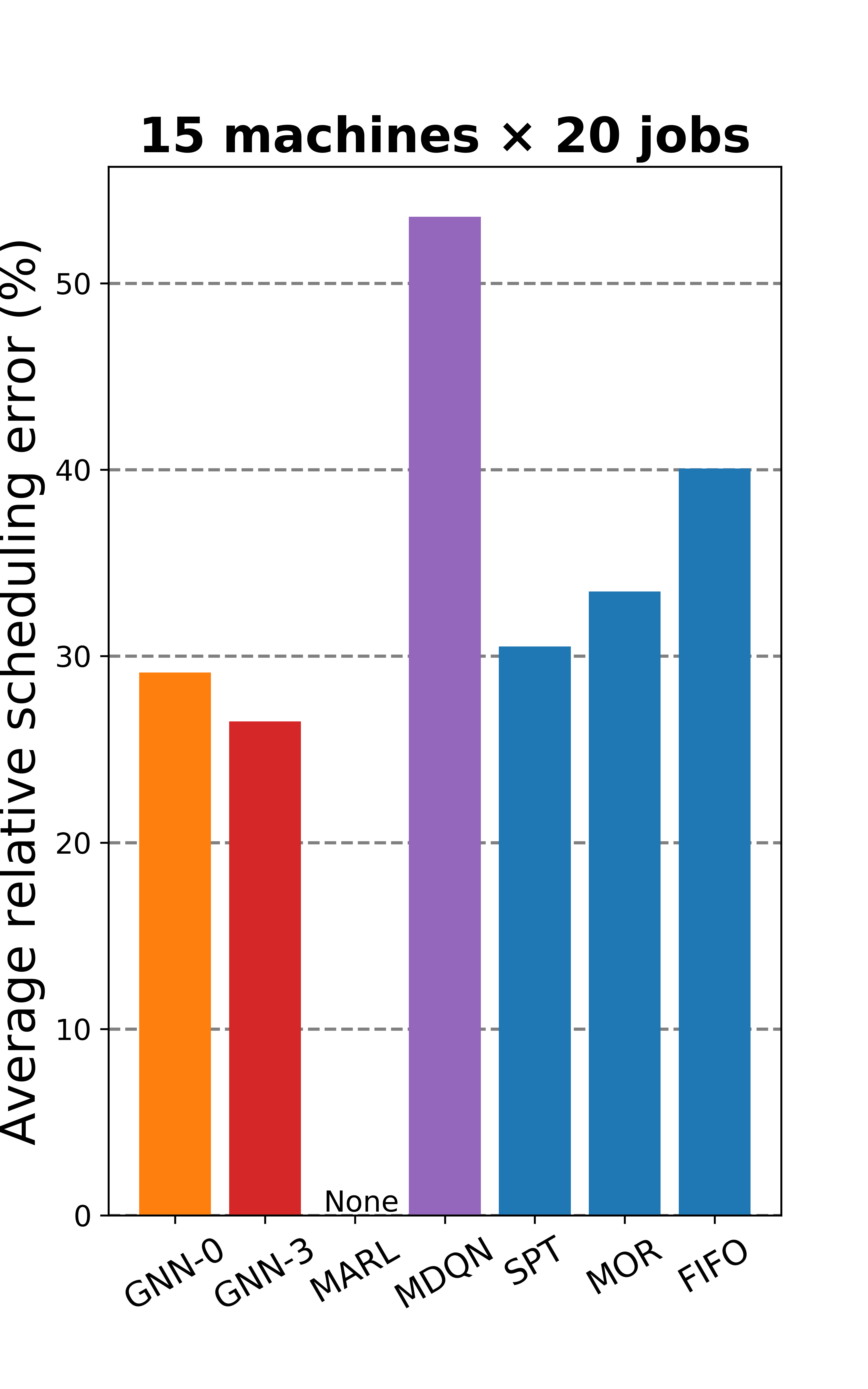}
\label{fig:subfig3}}
\caption{\textbf{Performance comparison with other RL-based JSSP scheduler}}
\label{fig:exp_3}
\end{figure}

Figure~\ref{fig:exp_3} shows that the proposed schedulers ($\mathrm{GNN}_1$, $\mathrm{GNN}_2$, and $\mathrm{GNN}_3$) gave the lowest errors for each test data set when the training and test data sets contained JSSP instances whose sizes were the same but whose processing times and order distributions were different. In particular, the GNN schedulers performed significantly better than the other learning models. Also, $\mathrm{GNN}_0$ trained with random JSSPs whose generating distributions were different from the test data set was still able to schedule all test JSSPs with slightly increased error compared with $\mathrm{GNN}_1$, $\mathrm{GNN}_2$, and $\mathrm{GNN}_3$. 
Based on the observation that the performance gaps between $\mathrm{GNN}_0$ and $\mathrm{GNN}_i$ ($i \in {1,2,3}$) are small, we can conclude that the node embedding scheme learned from small sized JSSPs can be generalized to solve large-sized JSSPs. This property is indeed desirable in that the GNN scheduler trained on a set of small JSSP instances can be directly used to schedule large JSSP instances that are impossible to solve with conventional models.

\section{Analysis on computational complexity}
We investigate the computational complexity of the proposed method for different sizes of JSSP instances. The number of computations for embedding node features typically increases as the size of the JSSP instance grows. To quantify this relationship, we use two factors, the number of machines, $m$, and the number of jobs, $n$, and investigate the impact of each factor on computational complexity. We conduct two folds of experiments: (1) fixing $m$ and varying $n$, (2) varying $m$, and fixing $n$. 

\begin{figure}[t]
  \centering
  \includegraphics[width=.75\textwidth]{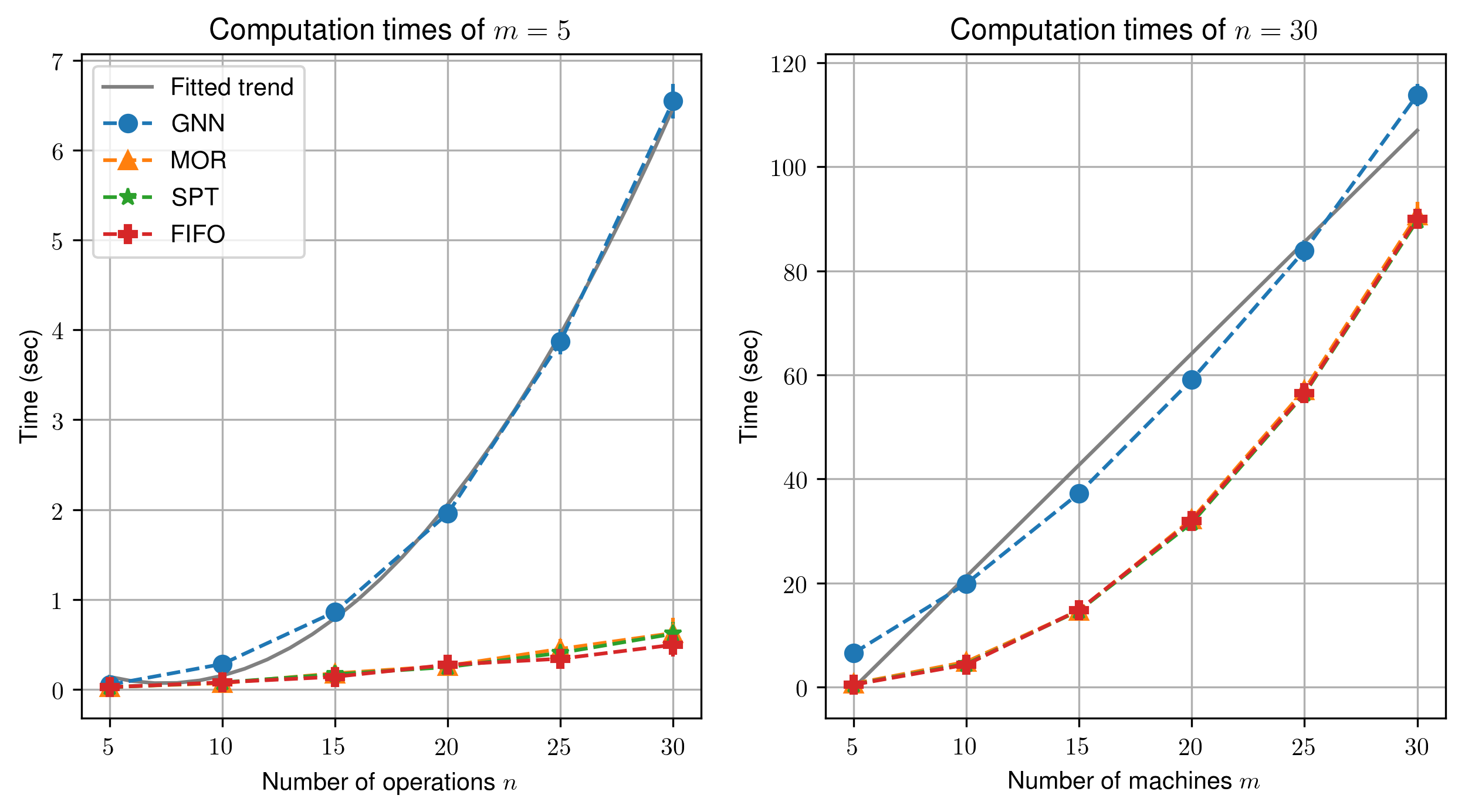}
  \caption{\textbf{computational times on varying instance sizes} [Left] 
  Computational times along the number of operations in JSSP graphs. [Right] Computational times along the number of jobs in JSSP graphs. We randomly generate 20 JSSP instances of each size and solves the problems with the proposed method to estimate the mean and standard deviation of computation times. The blue makers and ticks show the mean and standard derivation of computational times of the proposed GNN scheduler, respectively. Grey lines show fitted trends. Additionally, we compares the computation times of the three leading PDRs; MOR, SPT and FIFO. The orange, green, and red makers shows the mean computational times of MOR, SPT, and FIFO, respectively. We measure the computational times on a CPU machine, which is equipped with Intel(R) Core(TM) i9-7900X CPU @ 3.30GHz. (Best shows in color)}
  \label{fig:complexity_anlaysis}
\end{figure}

Figure~\ref{fig:complexity_anlaysis} show the result of the empirical complexity analysis. The left figure shows how the computational time increases as the number of jobs increases (while fixing the number of machines). In contrast, the right figure shows how the computation time increases with the number of machines. As the fitted second-order polynomial curve  $y=0.0125n^{2} - 0.1857n+0.7543$ with $R^{2}=0.9984$ represents, the computational time is likely proportional to the square of the number of jobs. Also, as the fitted first-order polynomial curve $y=4.2857m - 21.598$ with $R^{2}=0.9815$ indicates, the computational time seems to be proportional to the number of the machines. These results are supported by the fact that the number of edges in a JSSP graph is proportional to $n^2+nm$. 

As shown in Figure ~\ref{fig:complexity_anlaysis}, the proposed GNN-scheduler requires more computational time than PDRs. The GNN-scheduler needs to compute the node embedding by employ GNN before making the scheduling action, while PDRs do not compute any features. The computational time gaps generally increase with the number of operations (Figure~\ref{fig:complexity_anlaysis} [LEFT]) and decrease with the number of machines (Figure~\ref{fig:complexity_anlaysis} [RIGHT]). PDRs often conduct sorting operations to prioritize the possible scheduling actions and the computational cost of this operation increase with the size of JSSP instance. It is also noteworthy that our implementation of the GNN-scheduler does not employ any parallelization. We expect to effectively reduce the computational time by utilizing parallelization while computing node embeddings. Therefore, we can further decrease the computational time gap between PDRs and GNN scheduler in practice.

\section{Conclusion}
In this paper, we proposed a way to construct an efficient scheduler for job shop scheduling problems (JSSPs) with a GNN-RL framework. We formulated the sequential scheduling process of the JSSP as an MDP. Then, we proposed a graph representation for JSSPs, which serves as the state of the proposed MDP. We then employed a GNN to learn node embeddings that capture the spatial structure of the JSSP specified in the proposed graph. We proposed a scheduling module that utilizes the node embedding and recommends the operation to be processed. We used RL to train the parameters while considering the temporal property of the JSSP, i.e., the current time step's dispatching impacts the consequent dispatching. The proposed GNN scheduler learns the general properties of JSSPs rather than the properties of a specific JSSP instance. Therefore, the GNN scheduler can yield high-quality scheduling solutions for \textbf{any} JSSP without additional training.

The proposed GNN scheduler outperformed all PDRs when testing on JSSP instances generated by the distribution used for generating training instances. The proposed GNN scheduler showed robust performances on various sets of benchmark instances in contrast to PDRs, whose solution qualities fluctuated depending on the properties of the given JSSP. Also, the proposed GNN scheduler achieved better scheduling than the RL-based schedulers with the restriction on testing instances. Overall, the results demonstrate that our framework can produce a generalizable scheduling policy so that the GNN scheduler trained on only small JSSPs can generate high-quality schedules for large-sized JSSPs without additional training. 

Although the proposed GNN scheduler showed reliable performance on our testing set and benchmark JSSP instances, the computational time to compute the node embeddings rapidly increases as the size of JSSP increases. Because the total number of edges required to represent the JSSP state is proportional to $n^2+nm$, where $n$ and $m$ are the numbers of jobs and machines, respectively, the computational time to compute the node embeddings is also proportional to $n^2+nm$. In the future study, to decrease the computational time, we plan to introduce the new node type representing machines in the graph representation for the JSSP state. We can then construct disjunctive edges between the machine node and the job nodes of the JSSP graph, which can reduce the total number of edges from ${n \choose 2}$ to $ mn $. The reduced number of edges possibly reduce the total training and computational time. Lastly, the current formulation does not consider more practical constraints such as batch processes and waiting time regulations and flexible shop setup. Such formulations can be handled by expanding MDP formulation such that state transitions comply with the additional constraints. Furthermore, the reward function can be modified to cope with such extension. For instance, by adding the terminal reward that accounts for the mean tardiness or flow time on the proposed reward setting, the proposed RL-based approach can optimize the scheduling policy of different interests. In a future study, we will expand our MDP formulation to incorporate such complex constraints.

\bibliography{references}

\newpage
\appendix
\section{Generating distributions of benchmark JSSP instances}
\label{appendix:benchmark_jssp_gen_dists}
We summarize the generating procedure of each benchmark JSSP instance. To our best knowledge, The original manuscripts which proposed FT, LA, and SWV instances are not accessible.

\subsection{ORB instances}
The ORB instances are proposed by \cite{applegate1991computational}. The ORB instances are all size of $10 \times 10$. The authors collected five $10 \times 10$ instances from the different instances; MT10, ABZ5, ABZ6, LA19, LA20, and generated five their original instances named ORB1, ORB2, ORB3, ORB4, and OR5. They renamed the ten instances as ORBs \cite{ong2013hybrid}. The generation procedure of the originally proposed five instances ORB1-5 is not specified precisely. The authors describe the generation procedure as follows:
\begin{displayquote}
the problems ORB1 through ORB5 were generated in Bonn in 1986: the prescribed processing orders for each problem were generated by guests (with the challenge to make the problems "difficult" and for each problem, we generated two sets of processing times at random and retained that set which gave a problem having the greatest gap between the optimal value and a standard lower bound).
\end{displayquote}




\subsection{ABZ instances}
The ABZ5-9 instances are firstly proposed from \cite{adams1988shifting}. For all instances, the jobs have to be processed on all machines. The machine sequence of each job is randomly generated from a uniform distribution. The processing times of operations in ABZ5 are randomly drawn from the integer uniform distribution on the interval [50,100]. i.e., the processing times of ABZ5 $\sim U[50,100]$. The processing times of ABZ6, AB7, AB8 and AB9 follows $U[25,100]$, $U[11,40]$, $U[11,40]$ and $U[11,40]$, respectively.

\subsection{YN instances}
The YN1-4 instances are firstly proposed from \cite{yamada1992genetic}. The YN1-4 instances are all size of $20 \times 20$. They do not specify the generation scheme of machine sequences of jobs. The processing times of YN1-4 were sampled from $U[10,50]$.

\subsection{TA instances}
The TA01-80 instances are firstly proposed from \cite{taillard1993benchmarks}. The number of machines $m$ and jobs $n$ were drawn from $U[15,20]$ and $U[15,100]$, respectively. The processing times were drawn from $U[1,99]$. The machine orders were generated by the following sequence:

\begin{enumerate}
\item The operations are assigned to the machines without perturbation. i.e., the first machine processes the first operation.
\item From the first operation, swap the assigned machine with the operation that is randomly chosen from the succeeding operations. Repeat this procedure until the last operation.
\end{enumerate}

\end{document}